# Human–AI Construction of Bayesian Networks for Operational Decision Support: A Virtual Survey Approach

**Kumar Rahul**
*Decision Sciences and Operations Management Area*
*Indian Institute of Management Kozhikode, Kerala, India*
rahulk15phd@iimk.ac.in

**Shovan Chowdhury**[1]
*Decision Sciences and Operations Management Area*
*Indian Institute of Management Kozhikode, Kerala, India*
shovanc@iimk.ac.in

**Abstract**
Bayesian Belief Networks (BBNs) are powerful tools for decision-making under uncertainty. However, building their structures and estimating parameters are difficult. Currently, researchers must choose between relying on expert judgement or using large datasets to learn the structure and parameters of the network. We propose a new methodology using Large Language Models to bridge the gap between expert opinion and data-driven learning. This approach uses a panel of AI agents to estimate probabilities based on specific personas and context. We then apply a trimmed-mean rule to remove noise from these responses. We develop a six-step BBN framework and illustrate it to model "customer intention to consult a doctor in an alternative healthcare system". The model reveals that while self-efficacy appears to be a major factor, its actual causal impact is small. In contrast, subjective norms have a much stronger effect in modelling customers' intention. The most effective strategy is to improve both confidence and community norms simultaneously.

[1] Corresponding Author

## 1. Introduction

Operations research (OR) often confronts decision problems where uncertainty permeates every variable. Managers in healthcare, supply chain, and project settings face questions about adoption, disruption, demand, and outcomes. They need a framework that handles uncertainty while supporting transparent reasoning. Bayesian Belief Networks (BBNs) provide such a framework, combining graph theory and probability theory in a unified representation (Brady, 2020; Daphne & Nir, 2009). BBN adoption in operations research has nonetheless progressed slowly (Hosseini & Ivanov, 2020). Recent reviews trace this slow progress to one specific bottleneck, the elicitation of the conditional probability table (CPT) parameters (Guinhouya, 2023; Juliani & Maciel, 2024). Structure learning and parameter learning are two distinct steps to build a BBN (Heckerman, 1997). Both are demanding, and parameter learning is especially challenging when nodes have multiple parents.

For example, consider a node with two states and four parents, each with three states. Populating the CPT requires estimating 27 independent parameters. Few practitioners can elicit this many values from intuition, and the cognitive load of bounded rationality is well documented (Simon, 1955). Data-driven learning offers an alternative, but it requires substantial volumes of representative data that are difficult to assemble in management research. BBN literature in supply chain risk and project management largely relies on expert elicitation, which has known limitations regarding subjectivity and validation (Guinhouya, 2023; Hosseini & Ivanov, 2020). Data-driven approaches cluster in computer science and engineering, where data is plentiful (Juliani & Maciel, 2024). But core management domains lack a scalable middle path. OR researchers need a method to parameterize BBNs when real data is limited and expert time is expensive. Recent advances in large language models (LLMs) suggest a promising direction to fulfil this gap. LLMs can generate plausible probability estimates conditioned on context, simulate diverse personas, and respond to thousands of structured prompts at relatively low cost. None of these capabilities replaces empirical data, but they offer a fast and economical first pass at CPT specification.

Against this backdrop, our paper makes three contributions. We propose an LLM-assisted methodology for eliciting CPTs in Bayesian Belief Networks. The method uses a panel of synthetic agents, structured prompts, and a trimmed mean aggregation rule. It addresses the cognitive load problem in expert elicitation and the data-scarcity problem in machine learning approaches. Second, we integrate this methodology into a three-tier web application that combines a browser-based modelling dashboard with a Flask backend, the *pgmpy* Python library for inference, and provider-agnostic LLM adapters. The architecture combines a BBN structure specification, LLM-based parameter elicitation, forward sampling for synthetic data generation, maximum likelihood estimation for parameter refinement, and probabilistic and causal inference for decision support. It is a modular architecture that applies to any BBN, not only the example we present. Lastly, we present a numerical study of a health services decision problem with 8 variables and 99 independent parameters. The study tests 11 structural assumptions, identifies which interventions yield the largest gains in customer intention, and distinguishes causal effects from spurious correlations using the do-calculus. The results offer concrete guidance for operations managers in the alternative healthcare sector.

The remainder of the paper proceeds as follows. Section 2 reviews the foundations of Bayesian decision models, structure and parameter learning, and the role of LLMs in operations research. Section 3 presents the LLM-assisted methodology in detail. Section 4 describes the system architecture and workflow. Section 5 applies the methodology to the Ayurvedic consultation problem. Section 6 discusses validation and sensitivity analysis. Section 7 reflects on methodological implications, practical considerations, limitations, and directions for future research. Conclusion follows in Section 8.

## 2. Background and Literature Review

### 2.1 Bayesian Belief Network

A BBN is a directed acyclic graph (DAG) $G$ whose nodes are random variables and whose directed edges encode dependencies. The graph is paired with conditional probability tables that quantify each parent-child dependence. The graph and the CPTs together factorize the joint probability distribution by the chain rule (Daphne & Nir, 2009; Pearl, 1988). The $d$-separation criterion allows users to read conditional independence directly from the graph. Three reasoning patterns follow naturally, i.e., predictive (cause to effect), evidential (effect to cause), and intercausal (one cause given another). The do-calculus (Pearl, 1995) extends the framework to interventional queries of the form $P(Y \mid do(X))$ under specified graphical conditions. The combination makes BBNs attractive for decision support under uncertainty (Brady, 2020; Pearl & Mackenzie, 2018).

Constructing a BBN involves the steps of structure learning that identify the DAG, and parameter learning, which estimates the CPTs (Heckerman, 1997). As suggested by Zwirglmaier & Straub (2016), there are four approaches to learning the structure - transformation of existing probabilistic models; derivation from existing physical or empirical models; expert elicitation; and data-driven learning through constraint-based, score-based, or hybrid algorithms. However, each approach faces a parameter-learning challenge once the structure is fixed, since CPTs grow exponentially with parent count. Eliciting all values from a single expert is cognitively impractical due to bounded rationality (Simon, 1955) and estimating them from data requires representative observations at every parent configuration.

## 2.2 Application of BBN in Management Research

Several articles in the decision sciences literature use BBN in a variety of contexts. We organize this body of work along two dimensions. First, the review papers that map the field, and then the empirical applications that build and reason with specific networks.

Hosseini & Ivanov (2020) acknowledged that the application of BBN in supply chain risk management has progressed slowly. They considered the difficulty of generating BBNs via structure and parameter learning, both of which are primarily addressed through expert knowledge. Their review of 63 peer-reviewed articles from 2007 to 2019 nonetheless emphasized that BBN has transitioned from an emerging topic to a growing research area in supply chain resilience and risk modelling. Juliani & Maciel (2024) mapped 2,167 articles from Web of Science and Scopus across three decades. They showed that most empirical work concentrates on computer science and engineering rather than core management. Guinhouya (2023) systematically scanned 102 BBN studies in project management between 1999 and 2021. The applications cluster in construction and infrastructure projects (57), software and IT (21), engineering and manufacturing (13), and other domains (11). Sixty of the 102 articles originate from Asia (25 from China, 4 from India), compared with 22 from Europe and 15 from the USA. The author concluded that reliance on experts for parameter data makes the BBNs unreliable and weak in quality assessment.

From an empirical applications point of view, Nadkarni & Shenoy (2001) converted a qualitative input from a marketing expert into a fully specified Bayesian causal map. They used the network for probabilistic what-if and diagnostic reasoning about new-product launch decisions. The CPTs were also elicited directly from the expert. Forward and backward inference quantified how market dynamics, life-cycle stages, and leadership positions jointly affect aggressive launch strategies. Cinar & Kayakutlu (2010) extended causal cognitive mapping to national-level energy policy. They translated expert-derived causal maps of Turkey's energy system into a BBN that evaluates investment scenarios for renewable and nuclear energy. The expert group included academics in renewable energy, water resources, and bioenergy production, as well as political experts. They discretized historical data and refined conditional probabilities with expert judgment. Inference compared how alternative mixes of renewables and nuclear power influence greenhouse emissions and import dependency.

Ojha et al. (2018) designed a BBN to capture the propagation of disruptions across a multi-echelon supply chain. They built the structure using score-based learning (the K2 algorithm) on risk factors for suppliers, manufacturers, distributors, and retailers. CPTs combined simulation outputs with expert input. Inference identified the most vulnerable nodes and showed how simultaneous disruptions amplify system-wide performance losses. Hossain et al. (2020) modelled the interdependencies between port transportation and surrounding supply chains. Based on a review of 26 studies, the authors identified six factors driving port disruptions. CPTs were constructed using expert elicitation combined with Boolean operators. Sensitivity analysis revealed which factors most strongly drive disruption and logistics performance. In healthcare, Al Nuairi et al. (2024) developed a Tree Augmented Naive Bayes (TAN) model to identify provider-related drivers of patient experience. The authors used 434 observations from the British NHS Inpatient Survey conducted from 2017 to 2019. Continuous factor scores were discretized via *K*-means clustering. The TAN structure was learned automatically. Inference ranked factor importance and supported what-if analyses on improvements in trust, respect, or emotional support.

At the intersection of supply chain resilience and formal causal inference, Liu et al. (2022) formulated a causal Bayesian network over a multi-echelon chain. They used do-calculus to derive closed-form expressions for disruption probabilities under different supplier-intervention schemes. The expressions were embedded in a mixed integer nonlinear program, converting do-calculus into an optimization tool. Kabir et al. (2015) developed a BBN for water distribution networks, which are among the most expensive municipal assets. They identified 21 deterioration and consequence factors from the literature and expert opinion. A hybrid approach refined initial expert CPTs using data-driven EM learning from observed pipe, soil, hydraulic, and water-quality data. Inference computed pipe-level and network-level failure-risk indices. Duchessi & Lauría (2006) developed a BBN-based decision-support system for client-server IT implementation. Factor analysis on a national survey reduced 350 responses to 13 implementation factors and one benefit factor. A simulated annealing search with K2 scoring learned the structure. CPTs were estimated through MAP learning with Dirichlet priors. The fitted BBN supported what-if analysis on planning, project management, and user involvement to improve benefit realization. Yang & Bequette (2023) made the case for observational causal inference in manufacturing, where experiments are often impractical. They specified mechanistic structural causal models and used do-calculus to estimate intervention effects. The framework identified critical process parameters solely from observational data.

Two patterns emerge from this body of work. First, most studies rely on expert opinion to learn both the structure and CPTs of the BBN. Tree Augmented Naive Bayes and the K2 algorithm appear in a few studies, marking the data-driven exception rather than the rule. Second, of the studies reviewed, only Liu et al. (2022) and Yang & Bequette (2023) applied do-calculus for causal inference. Recently, Tafti & Shmueli (2025) argued that structural causal modelling and causal diagrams should be adopted more systematically in management research as complements to econometrics and potential-outcomes methods. They observed that DAGs make identification strategies, instrumental variable assumptions, and endogeneity issues more transparent for both authors and reviewers. We find this argument conclusive. However, BBN with do-calculus remains scarce in core management research. To the best of our knowledge, the integration proposed in this study has not been examined in the management science literature.

### 2.3 Generative AI in Management Research

The role of generative AI in management research is a new area shaping ongoing discussions. Carlson & Burbano (2026) argued the case for LLM use in social science research. They developed a foundational five-stage framework focused on data annotation tasks and examined how prompt engineering strategies shape model performance. Doshi et al. (2024) investigated generative AI for evaluating strategic alternatives by using AI-generated business models. Their findings suggest that generative AI often produces evaluations that are inconsistent and biased, but when aggregated, AI rankings tend to resemble those of human experts. Boussioux et al. (2024) explored "AI in the loop" for human-centred creative problem-solving, using strategic prompt engineering with GPT-4 to generate ideas on sustainable, circular-economy businesses. The authors' findings align with the previous study and conclude that outputs generated through a human-AI collaboration approach human-level creativity.

At the more specific intersection of LLMs and BBN construction, Babakov et al. (2024) proposed a methodology for structure elicitation. Several LLMs with diverse experiences are independently queried for candidate DAGs, and the final structure is obtained by majority voting. They demonstrate that LLMs can suggest plausible structures by drawing on implicit world knowledge. But they also warn that famous benchmark networks may have been memorized from training data, inflating apparent performance. However, the contribution targets the structure-learning aspect of BBN construction while leaving parameter learning open.

Beyond annotation, evaluation, and creative tasks, LLMs are beginning to enter BBN construction itself, but parameter learning has not been addressed in a similar way. CPT elicitation needs probability estimates conditioned on rich qualitative context, and LLMs excel at this kind of reasoning. They can simulate diverse personas and scale to thousands of structured prompts at relatively low cost, and aggregating responses across a panel of agents addresses individual-response bias (Doshi et al., 2024).

### 2.4 Positioning of the Present Study

We build on these threads to develop a methodology that integrates two ingredients. The first is an expert-based approach for structure learning, deriving the DAG from established theory rather than from data or from LLM-based elicitation. In our application case, the conceptual framework draws on social cognition models in health psychology, namely the Health Belief Model, Protection Motivation Theory, and the Theory of Planned Behavior (Ajzen, 1991; Conner & Norman, 2015; Janz & Becker, 1984; Norman et al., 2015). This framework can be instantiated to explore any specific health behaviour that is of interest to the service provider.

The second ingredient is LLM-based virtual surveys for parameter learning. We collaborate with LLMs through strategic prompt engineering, guiding each model to explore the search space under carefully crafted instructions. We aggregate responses across a panel of synthetic agents using a trimmed mean, following the empirical finding of Doshi et al. (2024) that aggregated LLM responses resemble human expert rankings. The aggregated values are used to populate the initial CPT, after which forward sampling and maximum likelihood estimation further refine them.

To the best of our knowledge, this combination has not previously been explored in the OR and management science literature. The proposed approach addresses two important challenges: the cognitive burden associated with traditional expert elicitation and the limited data availability that often constrains purely machine-learning-based methods. At the same time, it preserves the causal interpretability offered by BBNs. By integrating social cognition theory for network construction, generative AI for CPT elicitation, and BBN-based causal inference, this study introduces a novel methodological framework that may open new avenues for customer-intention modelling and broader operational decision-making applications.

## 3. Methodology: LLM-Assisted Construction of Bayesian Decision Model

### 3.1 Problem Setting and Notation

We consider a generic operational decision problem in which the decision-maker faces uncertainty over a set of discrete random variables. Some variables represent controllable levers, others capture contextual conditions beyond managerial control, and one or more correspond to the outcome of interest. Let $X = \{X_1, X_2, \ldots, X_n\}$ denote the set of variables, and let $Y \subset X$ represent the decision-relevant outcome nodes. Depending on the operational context, $Y$ may correspond to outcomes such as adoption, failure, demand, disruption, or behavioral intention.

Let $G = (V, E)$ denote a DAG defined over these variables. Each variable $X_i$ takes values from a finite space $Val(X_i)$ with cardinality $k_i$. The graph specifies the dependence structure among variables, allowing the joint probability distribution to be factorized according to the Bayesian network chain rule:

$$P(X_1, \ldots, X_n) = \prod P(X_i \mid Parents(X_i)) \ \forall\, i.$$

Each conditional distribution $P(X_i \mid Parents(X_i))$ is represented through a CPT, which specifies the probability of every state of $X_i$ for all possible combinations of parent states.

The modeling problem can be understood through three parameterization regimes. At one extreme, the full joint distribution requires $(\Pi\, k_i) - 1$ free parameters, where $\Pi\, k_i$ represents the size of the joint state space. This quantity grows exponentially with the number of variables, making direct specification infeasible in most practical settings. At the other extreme, a naive independence model with a parametric link function, such as logistic regression, substantially reduces the parameter count but cannot adequately represent multidirectional dependencies or probabilistic reasoning among interconnected variables. The factorized BBN provides an intermediate representation between these two extremes. For a node $X_i$, the corresponding CPT requires $(k_i - 1) \times \Pi\, k_j$ parameters, where $i$ indexes the node and $j$ runs over its parents. Summing across nodes gives the total parameter count for the BBN. Summing across all nodes gives the total number of parameters required for the BBN. This factorization substantially reduces dimensionality while preserving the ability to model conditional dependence structures and multidirectional inference.

The factorization provides important modeling flexibility. Adding new states to a node affects only the CPT of that node and the CPTs of its child nodes, while the remainder of the network remains unchanged. Although this modularity simplifies structural modification, the challenge of parameter elicitation remains significant. Even moderately sized networks with multiple parent nodes may require a large number of independent parameter estimates for each CPT. The central issue addressed in this study is how to estimate these parameters when neither sufficient expert time nor representative datasets are available.

### 3.2 Methodology Workflow

The proposed methodology proceeds in six steps. Figure 1 summarizes the flow and identifies which actor performs each step.

**Figure 1:** *Six-step process for BBN construction*

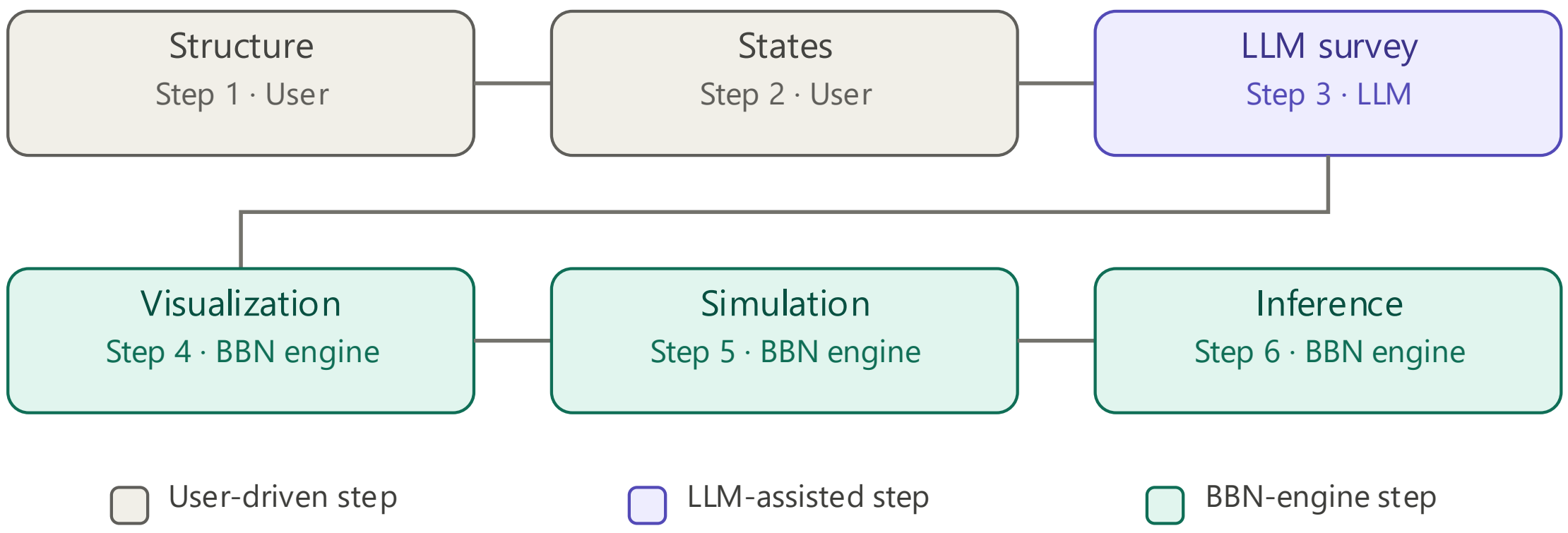

The first step defines the network structure, in which the user specifies nodes, edges, and a textual description for each variable. The structure builds from theory, a literature-based thought experiment, or a domain expert. We do not learn structure from data in the proposed method, because the data-scarcity problem also affects structure learning. Discrete states for each node are specified in Step 2, along with state counts and labels that either follow a scale-development practice or are based on established literature. The user also writes a semantic description for each state, which is used in the LLM prompts in the next step. The methodological core is Step 3, which uses an LLM-based virtual survey. It elicits the CPTs by generating a panel of synthetic agents, which are then queried to obtain the probability distribution at each parent configuration for each node. The results are aggregated to produce an initial CPT, as detailed in Sections 3.3 and 3.4. Once the CPTs are populated, the system visualizes the network so the user can inspect the structure and the initial values. Manual override is supported, and if a particular CPT row looks implausible, the user can edit the values before proceeding. In simulation step 5, forward sampling generates synthetic data from the network, and maximum likelihood estimation refines the CPTs based on the data. The refinement smooths sampling noise and produces a self-consistent set of inference parameters. Inference is the final step, supporting probabilistic queries through variable elimination and causal queries through truncated factorization. These queries answer the decision-maker's questions about the network. The handoff between LLM and the BBN engine occurs at the end of Step 3. After Step 3, the BBN engine takes over, while the user remains in the loop throughout, particularly for visualization and query specification.

### 3.3 LLM-Assisted Virtual Survey Design

The virtual survey is designed to replicate the structure of a conventional consumer or expert survey. It consists of three key components: agent personas, structured prompts, and quality-control mechanisms. Figure 2 illustrates the overall workflow. Four input elements - persona descriptions, network context, node states, and node descriptions are integrated into a standardized prompt template. The LLM-based survey is then executed for each agent across all parent-state configurations. Finally, a trimmed-mean aggregation procedure is applied to generate the initial CPT for each node.

**Figure 2:** *CPT generation using virtual survey*

Persona
description
Network
context
Node
states
Node
descriptions
Prompt template
Four components combined per query
LLM survey
Per agent, per parent configuration
Trimmed mean aggregation
Drop top and bottom 10%, normalize
Initial CPT for node
P(Xi | Parents = p)
User-defined input
LLM-assisted step
Output parameter

The persona generator creates a panel of N synthetic agents, where N is decided by the user. A single Application Programming Interface (API) call to the LLM returns N profiles, each encoding demographics relevant to the decision context. The persona prompt, therefore, includes the network description and the geographic or cultural context. This step grounds the synthetic responses in plausible local conditions. Demographics may include age, gender, education, income, occupation, and any other dimension the user deems necessary. Personality diversity is essential; without it, the panel collapses to a single representative response and loses the variance that makes aggregation meaningful.

The next element is the prompt template. Each node receives its own prompt, with root nodes (no parents) receiving a prior-elicitation prompt that asks the LLM to estimate $P(X_i)$ given the persona, the node states, the node description, and the network context. Conditional nodes (with one or more parents) receive a different prompt for every combination of parent states, asking the LLM to estimate $P(X_i | Parents(X_i) = p)$ for each specific parent configuration $p$. Every prompt combines four input components and a query. The persona description sets the respondent's identity. One of the N synthetic agents is bound to each call. The network context then positions the node within the larger decision problem and lists the network edges relevant to the focal node. The node states list the discrete labels the variable can take, for example, Low, Medium, and High. Finally, the node descriptions provide the semantic meaning of the variable and define what each state implies in the application domain.

The query then asks for a probability vector that sums to 1 and specifies the output format. The four components are not interchangeable. Node states provide the LLM with a discrete set of choices. Node descriptions tell the LLM what each state means. Network context anchors the LLM in the broader dependence structure. Personality description sets the perspective from which probabilities are estimated. Removing any one of the four degrades response quality.

Quality controls operate at two levels. At the prompt level, we include explicit format instructions and an example output format. For conditional nodes, the conditional prompt anchors the LLM on the dependence structure. At the aggregation level, we use a trimmed mean, as discussed in the next subsection.

**3.4 From Agent Responses to CPTs**

Each N agent returns a probability vector for each cell of each CPT. For a given node $X_i$ and parent configuration $p$, we have $N$ estimates of $P(X_i \mid Parents(X_i) = p)$. We aggregate these $N$ estimates into a single distribution. We use a trimmed mean to exclude the top 10% and bottom 10% of estimates and take the mean of the remaining 80%. The trimmed mean mitigates the effects of outliers, since LLM responses occasionally exhibit hallucinations or extreme values that sharply diverge from the panel consensus. The rule reduces their influence without discarding the bulk of the responses.

After trimming, we normalize the vector so that the probabilities sum exactly to 1, since floating-point arithmetic and the trimming step can introduce small deviations. Normalization ensures that the CPT remains a valid probability distribution. The procedure handles missing or noisy responses gracefully. If an agent returns an invalid response (non-numeric, missing states, sum far from 1), we discard that response and proceed with the remaining agents. As long as the post-trimming sample size remains reasonable, the aggregation is robust.

**3.5 Parameter complexity and computational cost**

The factorization of the joint distribution dramatically reduces the parameter count. For $n$ discrete variables with cardinalities $k_i, i = 1, 2, \ldots, n$, the full joint distribution requires $(\prod_i k_i) - 1$ independent parameters. In contrast, a factorized BBN requires $\sum_i (k_i - 1) \prod_{j \in Pa_i} k_j$ parameters, where $Pa_i$ denotes the parent set of variable $X_i$. The number of LLM API calls is a related but distinct quantity. Each parent configuration of each conditional node requires one call per agent. Each root node requires one call per agent for its prior. Persona generation requires one additional call. The total call count is $n_{API} = 1 + N \times \left(|R| + \sum_{i \in C} \prod_{j \in Pa_i} k_j\right)$, where $R$ is the set of root nodes and $C$ is the set of conditional nodes. This formula has practical implications. Total cost scales linearly with the panel size $N$, and with the number of parent configurations across the network. Adding parents to a node grows the configuration count multiplicatively.

Time to completion of the survey depends on concurrency: higher concurrency completes the survey faster, but increases the risk of rate limiting from the LLM provider. The choice of model and concurrency level should follow the user's budget and time constraints. Frontier models cost more per call, while smaller balanced models cost less and often suffice for CPT elicitation, since the trimmed mean over a panel mitigates individual noise. We provision caching to limit how many times the survey is triggered for the same configuration. It reduces redundant work by computing a hash key from the persona description, network context, node descriptions, and query template. Subsequent runs with the same configuration retrieve cached results, whereas any variation in the input triggers a fresh survey.

### 3.6 Integration with BBN inference

The trimmed-mean CPTs serve as the initial parameters for the BBN, and downstream inference proceeds in three stages: forward sampling, parameter refinement, and query answering. Forward sampling generates synthetic data from the network. This is achieved by ordering the nodes topologically and sampling each conditional node based on its already-sampled parents. The resulting synthetic dataset, by construction, follows the joint distribution of the network. These synthetic observations are generated from the Bayesian network through forward sampling. As the number of observations increases, empirical frequencies converge to the probabilities encoded in the network. The resulting synthetic dataset is then used to estimate conditional probability tables via maximum likelihood estimation, yielding data-driven parameter estimates that approximate the original elicited probabilities. These MLE-derived CPTs replace the initial trimmed-mean estimates. The refinement smooths sampling noise and produces a self-consistent set of parameters.

Query answering completes the pipeline through different machinery for probabilistic and causal queries. Probabilistic inference computes posterior probabilities given observed evidence using the variable elimination method ("Bayesian Network Representation," 2009). Causal inference instead requires graph mutilation where the truncated factorization severs incoming edges to the intervention node, fixes that node to a constant, and runs variable elimination on the mutilated graph ("Causality," 2009; Pearl, 2000). The procedure is the operational form of do-calculus – "rule 2" also known as the back-door criterion (Pearl, 2000). It returns $P(Y \mid do(X))$ for any set of intervention nodes X and the contrast between $P(Y \mid X)$ and $P(Y \mid do(X))$ reveals confounding effects in the network.

The methodology supports decision queries of three types: predictive ("what happens if we observe condition X?"), diagnostic ("what condition explains observed outcome Y?"), and interventional ("what happens if we force condition X?"). The combination provides the decision-maker with insights for every possible granular combination.

## 4. System design for decision support

### 4.1 Architecture

We have implemented the methodology using a three-tier architecture, as shown in Figure 3, that separates concerns and supports modularity. The frontend tier runs in the browser and provides a modular dashboard with persistent navigation, interactive Data-Driven Documents (D3.js) network visualizations and responsive grid layouts for CPT inspection. The interface lets the user define a network, view its structure, examine CPTs, run simulations, and pose queries. The backend tier runs on a Flask server that exposes a Representational State Transfer (REST) API consumed by the frontend. The backend orchestrates four services: a simulation service that handles forward sampling and parameter refinement; an LLM interface that routes virtual-survey requests to the appropriate model provider; a configuration manager that persists network definitions, state descriptions, and prompt templates; and a caching layer that prevents redundant LLM calls.

The external tier comprises two service families. One is *pgmpy*, an open-source Python library for probabilistic graphical models (Ankan & Textor, 2024) which handles forward sampling, MLE-based parameter learning, variable elimination, and graph mutilation for causal queries. The other is the family of commercial LLM providers, including OpenAI, Anthropic, and Google. The LLM interface abstracts away provider differences, allowing the user to swap models without code changes. The architecture is modular, and thus adding a new LLM provider requires only implementing a corresponding adapter in the LLM interface layer. Similarly, replacing *pgmpy* with another graphical model library requires modification only within the simulation service via a dedicated adapter. In both cases, the frontend layer remains unaffected. The application is currently hosted privately on *Hugging Face* and may be shared upon request.

**Figure 3:** *Three-tier architecture for system design*

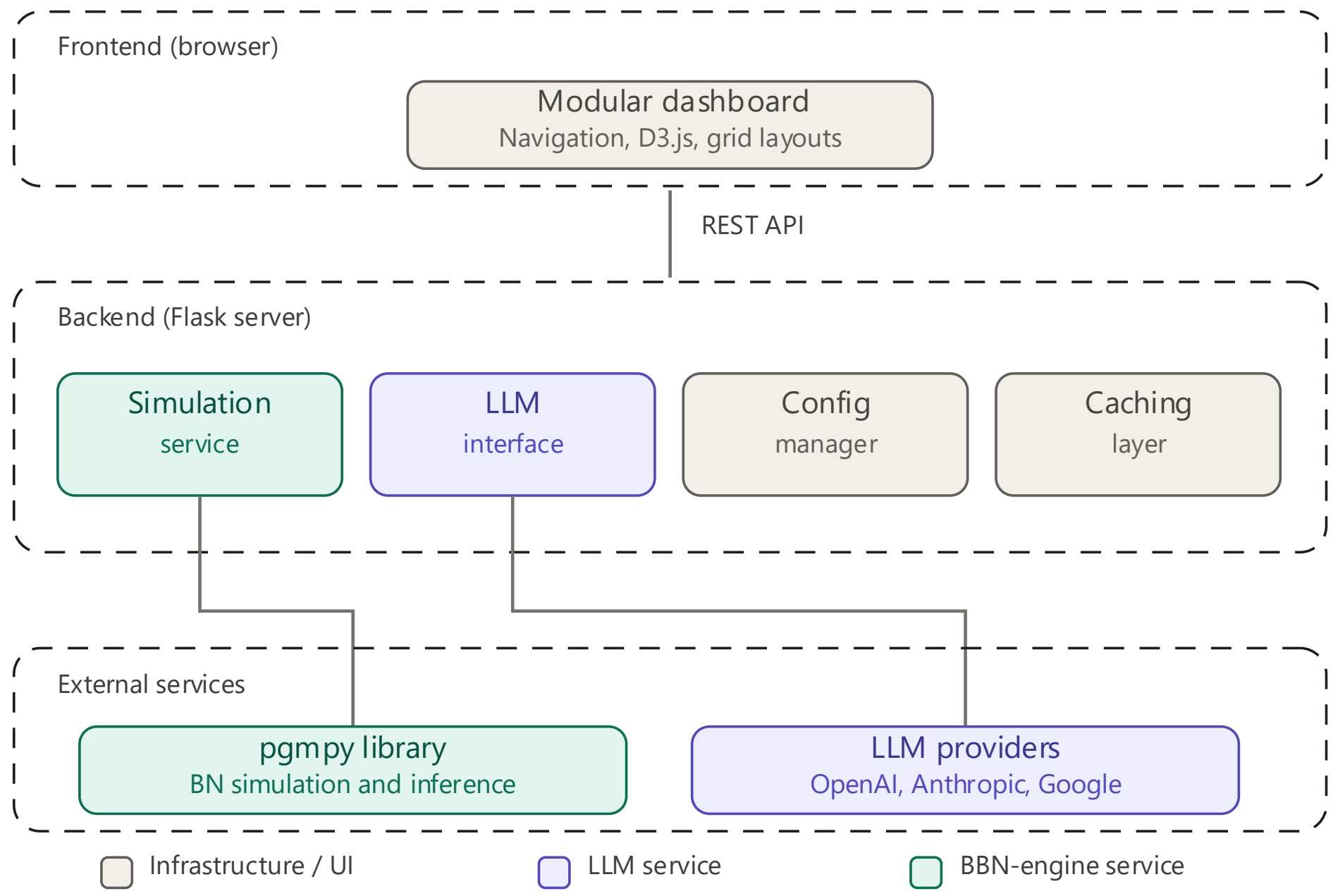

## 5. Application to Health-Service Decision Problem

### 5.1 Operational Context and Decision Problem

Alternative healthcare in India represents a substantial but underserved segment of the broader healthcare system. Ayurveda and naturopathy compete with Western medicine for a share of customer attention and spending. Customer perceptions in this segment are shaped by societal biases, competing marketing, and the desire to retain a measure of personal control over health choices (Rajamma & Pelton, 2010). Providers in this segment also face operational challenges with demand being volatile, and resource-allocation decisions, including practitioner deployment, treatment program design, and outreach campaigns, depend on understanding which customer segments are most likely to engage.

We focus on lifestyle diseases as the application context, with diabetes as a prevalent and well-defined example. The customer faces a meaningful choice between conventional Western medicine and Ayurvedic consultation, and many use both. The provider's decision is how to identify and reach customers whose intention to consult an Ayurvedic doctor is high, or whose intention can be moved through targeted intervention. The decision-maker has finite resources and competing levers. Should the provider invest in initiatives that reduce perceived barriers, in community endorsement campaigns that strengthen subjective norms, or in customer education programs that enhance self-efficacy? Each lever has a different cost and a different expected lift in customer intention. The decision-maker needs a model that quantifies these effects under uncertainty.

Traditional analytical methods, such as regression and structural equation modelling, assume linearity and offer limited support for clean causal queries. Both approaches require survey data volumes that a small alternative healthcare provider is unlikely to collect. BBN addresses these gaps by handling non-linear interactions, supporting causal reasoning through do-calculus, and remaining tractable at a moderate scale. The challenge is the CPT specification, which our LLM-assisted methodology resolves.

### 5.2 BBN Structure and Variables

The network structure is grounded in established social cognition theory. The health belief model (HBM), protection motivation theory (PMT), and the theory of planned behavior (TPB) provide the constructs (Ajzen, 1991; Conner & Norman, 2015; Janz & Becker, 1984; Norman et al., 2015). The network structure is specified using an expert-informed approach grounded in social cognition theory. The network comprises eight variables, of which three are root nodes representing exogenous customer characteristics. Perceived Barriers (PBa) capture the practical and psychological cost of consulting and following an Ayurvedic prescription. We introduce a novel construct, Perceived Barriers of Competition (PBaC), that captures the same construct for a dominant alternative product or service competing with the focal product or service. In the context of this instantiation, we define “western medicine” known as “conventional medicine” as a dominant alternative to "ayurvedic medicine” also known as non-conventional medicine . Health Status (HS) captures the customer’s self-perceived physical and mental well-being.

Four variables act as mediators. Attitude (A) is a learned disposition toward Ayurvedic consultation, formed through evaluation of barriers and benefits relative to the customer’s health condition. Health Consciousness (HC) is the extent to which health concerns are integrated into daily life. Self-Efficacy (SE) is the customer’s belief in their capability to follow an Ayurvedic regimen. Subjective Norms (SN) capture the perceived social pressure from significant others (Bandura, 1977; Jayanti & Burns, 1998). The eighth variable, Intention (I), represents the customer's likelihood or propensity of consulting an Ayurvedic physician for the management of diabetes. Bayesian Belief Network structure for the customer-intention model is demonstrated in Figure 4.

**Figure 4:** *Bayesian Belief Network structure for the customer-intention model. Three root nodes (PBa, PBaC, HS) feed Attitude. HS also feeds Health Consciousness. Attitude and HC feed Self-Efficacy. SE feeds Subjective Norms. Attitude, HC, SE, and SN all feed Intention.*

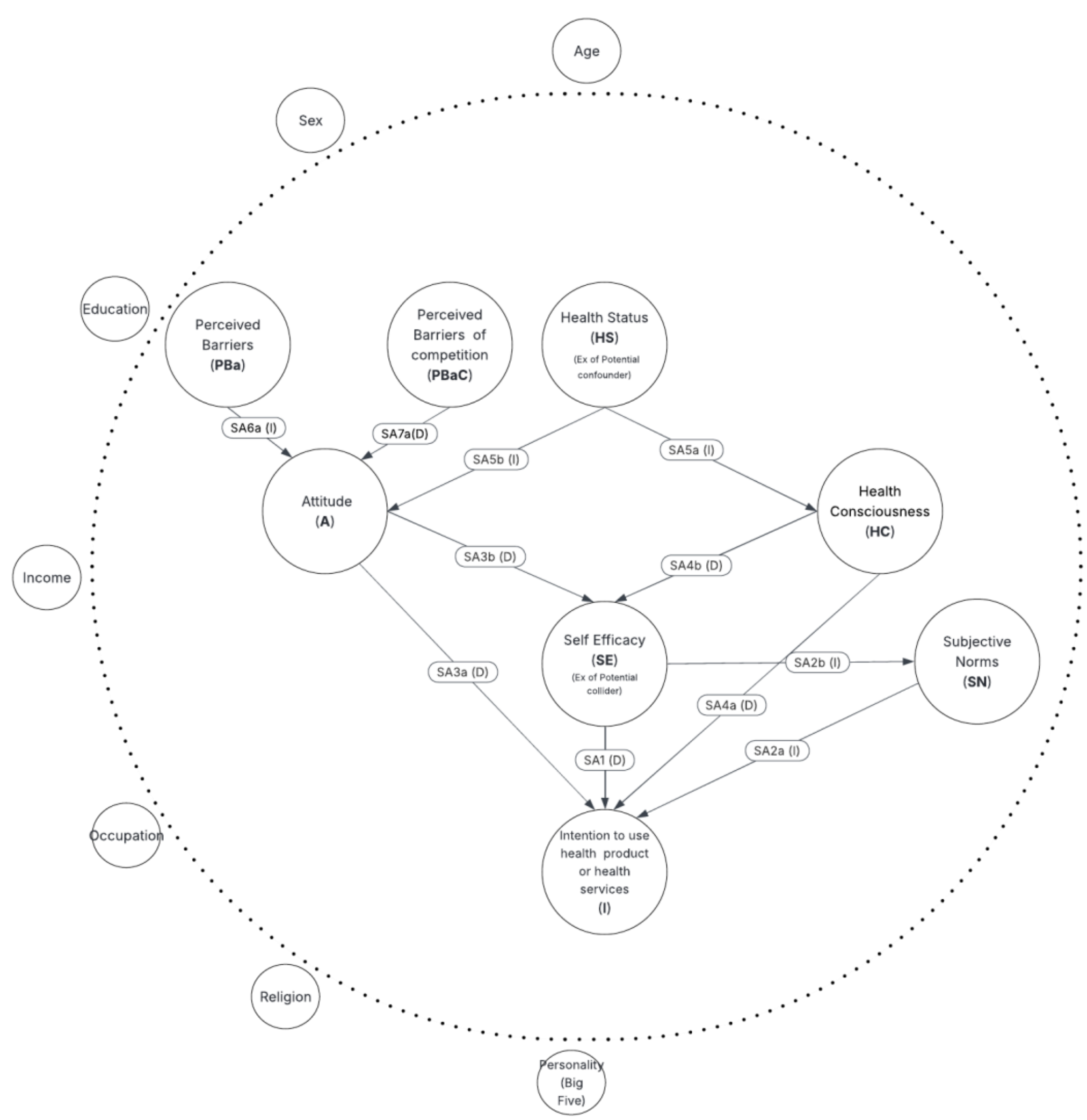

The discrete states for each variable follow scale development practices. PBa, PBaC, HC, SE, and SN each have three states. HS, A, and I each have two states. The state vector is (3, 3, 2, 2, 3, 3, 3, 2). Specific labels include Low, Medium, and High for PBa and PBaC; Bad and Good for HS; Low and High for Attitude. Low, Moderate, and High for HC; Low, Average, and High Control for SE. SN takes the state of Not Influenced, Somewhat Influenced, and Highly Influenced for SN. The intention to consult an Ayurvedic doctor (I) takes on the state values of Low and High. Eleven structural assumptions in Table 1 emerge from the social cognition literature and translate directly into testable propositions for the BBN.

**Table 1:** *Structural Assumptions to explore intention to consult an Ayurvedic doctor for treatment of diabetes.*

| Structural Assumption | Description |
|---|---|
| SA1 | Individuals with higher self-efficacy (SE) will have a higher intention (I) to use alternative healthcare products or services. |
| SA2a | Individuals under stronger subjective norm (SN) influence will show lower intention (I) to use alternative healthcare products or services. |
| SA2b | Individuals with lower self-efficacy (SE) will be highly influenced by subjective norm (SN) and will show lower intention (I) to use alternative healthcare products or services. |
| SA3a | Individuals having a favourable attitude (A) towards an alternative healthcare product or service will exhibit a higher intention (I) to use the alternative healthcare product or service. |
| SA3b | Individuals having a favourable attitude (A) towards alternative healthcare products and services and higher self-efficacy (SE) will show higher intention (I) to use alternative healthcare products or services. |
| SA4a | Individuals who perceive a higher level of health consciousness (HC) will show higher intention (I) to use alternative healthcare products or services. |
| SA4b | Individuals who perceive a higher level of health consciousness (HC) and exhibit higher self-efficacy (SE) will show higher intention (I) to use alternative healthcare products or services. |
| SA5a | Lower perceived health status (HS) and higher health consciousness (HC) will be associated with a higher intention (I) to use alternative healthcare products or services. |
| SA5b | Lower perceived health status (HS) will lead to a more favourable attitude (A) toward the intention to use alternative healthcare products or services. |
| SA6a | Individuals with a higher perceived barrier (PBa) to alternative health products or services will show an unfavourable attitude (A) toward their intention to use them. |
| SA7a | Individuals with a higher level of perceived barriers to the dominant alternative (PBaC) will have a more favourable attitude (A) toward their intention to use alternative healthcare products or services. |

## 5.3 LLM-elicited CPTs for the health model

The implementation used Gemini 2.5 Flash as the LLM, with a panel of N = 100 synthetic agents. The persona description prompted the LLM to generate diverse profiles of Indian customers who consult both an ayurvedic doctor and a general physician for lifestyle diseases. The prompt specified variation across age (21-30, 31-40, 41-50, 50+), sex, education, income, occupation, and religion. A single API call returned 100 distinct profiles. Sample profiles included a 48-year-old housewife from Uttar Pradesh, a 32-year-old software engineer from Bengaluru, and a 41-year-old university professor from Hyderabad. The profiles capture the demographic mix an Indian alternative healthcare provider would expect to encounter.

The API call followed the formula described in Section 3.5. Persona generation used 1 call. The three root nodes required 100 calls each, for a total of 300. The five conditional nodes require calls equal to the parent configuration count multiplied by the panel size. Attitude with three parents (PBa, PBaC, HS) required 18 × 100 = 1,800 calls. Health Consciousness with one parent (HS) required 2 × 100 = 200 calls. Self-Efficacy with two parents (A, HC) required 6 × 100 = 600 calls. Subjective Norms with one parent (SE) required 3 × 100 = 300 calls. Intention with four parents (A, HC, SE, SN) required 54 × 100 = 5,400 calls. Conditional nodes totalled 8,300 calls. The grand total was 8,601 calls.

The full survey ran in approximately 120 minutes at 20 concurrent calls. Per-call token consumption averaged 800 input and 50 output tokens, and the total cost was approximately USD 6 (INR 495) at the chosen model (subject to change). We implemented persistence via caching, which prevented re-execution. The hash of the persona description, network context, and node descriptions was used to pull the cached results.

Each CPT cell received 100 estimates from the panel. We aggregated using the trimmed mean rule. The top 10% and bottom 10% of estimates were excluded. The mean of the remaining 80% was normalized to sum to 1.0. The aggregated values populated the initial CPTs. Manual review of the CPTs found the values qualitatively plausible, with no extreme corrections required.

### 5.4 Simulation design and inference questions

The initial CPTs served as input to the forward-sampling simulation. We generated 100,000 synthetic samples using the *pgmpy* package (Ankan & Textor, 2024). The sample size was chosen to ensure that the empirical statistics converge to the true network probabilities. The MLE-derived CPTs replaced the initial trimmed-mean estimates and served as the anchor for all subsequent inference. We show the learnt CPT in Appendix A.

The inference questions followed the eleven structural assumptions developed from the social cognition literature. Each assumption translates into a specific probabilistic query. For example, the assumption that higher self-efficacy increases the intention to use alternative healthcare maps to comparing $P(I = High \mid SE = Low\ Control)$, $P(I = High \mid SE = Average\ Control)$, and $P(I = High \mid SE = High\ Control)$ against the baseline marginal $P(I = High)$.

We complemented the predictive queries with causal queries on the most actionable levers, including Subjective Norms, Self-Efficacy, Attitude, and Health Consciousness. For each lever X, we computed $P(I = High \mid X = high\ state)$ as the observational query and $P(I = High \mid do(X = high\ state))$ as the interventional query. The contrast between the two reveals confounding, as suggested by Pearl (2000), do-calculus. We also examined joint interventions. For example, intervening simultaneously on SE and SN tests the combined effect of building customer confidence and shifting community norms.

### 5.5 Results: Decision-Relevant Insights

The marginal probability of high intention in the simulated population is $P(I = High) = 0.534$. In the absence of additional evidence, about 53% of the population has a high intention to consult an Ayurvedic doctor for diabetes, and this baseline serves as the benchmark for all subsequent comparisons.

Self-Efficacy is a strong observational predictor: $P(I = High \mid SE = Low\ Control) = 0.277$ and $P(I = High \mid SE = High\ Control) = 0.837$. Compared to the baseline of 0.534, this suggests that customers with strong self-efficacy are far more likely to intend to have an Ayurvedic consultation.

The causal picture differs sharply. The interventional query $P(I = High \mid do(SE = High\ Control))$ returns 0.544, giving a causal lift of −29.3% ($defined\ as\ P(Y \mid do(X)) - P(Y \mid X)$). The negative lift implies that the observed correlation overstates the causal power of SE, with confounding running through Attitude. Customers who develop high SE typically also hold a favorable attitude, so forcing SE high without also shifting Attitude removes this confounded boost. The true causal effect, defined as $P(Y \mid do(X)) - P(Y)$, is just +0.9%, meaning the intervention on SE alone yields only a small gain.

Subjective Norms show the opposite pattern. The observational query $P(I = High \mid SN = Highly\ Influenced) = 0.430$ sits below the baseline, while the interventional query $P(I = High \mid do(SN = Highly\ Influenced)) = 0.609$ yields a causal lift of +17.9%. The positive lift implies that confounders suppress the observed effect, and the true causal effect of +7.5% is meaningful. An intervention that shifts community norms toward favoring Ayurvedic consultation produces a real and substantial gain in customer intention.

Joint interventions amplify the effect further. Forcing SE to High Control and SN to Highly Influenced together yields a true causal effect of approximately +15%, while forcing HC to High and HS to Good yields +9.6%. By comparison, SN alone yields +7.5%, and SE alone yields +0.9%. The ranking provides clear guidance for resource allocation: the largest gains come from joint interventions on customer confidence and subjective norms, while the smallest single-lever gain comes from self-efficacy alone, despite its strong observational signal.

The eleven structural assumptions yielded mixed verdicts. Seven assumptions were validated, with the model's behavior matching directional predictions from the social cognition literature. Three assumptions were not validated. From Figure 4, SA3a posited that a favourable Attitude alone leads to higher Intention, but the data showed otherwise, suggesting that Attitude needs to be combined with strong SE or supportive SN to drive Intention. SA3b posited a similar joint effect of Attitude and SE and was rejected for related reasons. SA7a posited that higher perceived barriers of the Western alternative lead to a favourable attitude toward Ayurveda, but the data did not support this, suggesting that the customer's attitude is shaped more by the absolute properties of Ayurveda than by the relative weakness of Western medicine. One assumption, SA5a, was inconclusive. The directional pattern was present, but the magnitude was small. Table 2 illustrates the result.

**Table 2**

*Results on the Structural Assumptions to explore Intention to consult an Ayurvedic doctor for treatment of diabetes.*

| Structural Assumption | Description | Result |
|---|---|---|
| SA1 | Individuals with higher levels of self-efficacy (SE) will show higher intention (I) to use alternative healthcare products or services. | Validated |
| SA1 | Individuals with higher self-efficacy (SE) will have a higher intention (I) to use alternative healthcare products or services. | Validated |
| SA2a | Individuals under stronger subjective norm (SN) influence will show lower intention (I) to use alternative healthcare products or services. | Validated |
| SA2b | Individuals with lower self-efficacy (SE) will be highly influenced by subjective norm (SN) and will show lower intention (I) to use alternative healthcare products or services. | Not Validated |
| SA3a | Individuals having a favourable attitude (A) towards an alternative healthcare product or service will exhibit a higher intention (I) to use the alternative healthcare product or service. | Not Validated |
| SA3b | Individuals having a favourable attitude (A) towards alternative healthcare products and services and higher self-efficacy (SE) will show higher intention (I) to use alternative healthcare products or services. | Validated |
| SA4a | Individuals who perceive a higher level of health consciousness (HC) will show higher intention (I) to use alternative healthcare products or services. | Validated |

| Structural Assumption | Description | Result |
|---|---|---|
| SA4b | Individuals who perceive a higher level of health consciousness (HC) and exhibit higher self-efficacy (SE) will show higher intention (I) to use alternative healthcare products or services. | Inconclusive |
| SA5a | Lower perceived health status (HS) and higher health consciousness (HC) will be associated with a higher intention (I) to use alternative healthcare products or services. | Validated |
| SA5b | Lower perceived health status (HS) will lead to a more favourable attitude (A) toward the intention to use alternative healthcare products or services. | Validated |
| SA6a | Individuals with a higher level of perceived barriers to the dominant alternative (PBaC) will have a more favourable attitude (A) toward their intention to use alternative healthcare products or services. | Not Validated |

The decision-relevant takeaways for an Ayurvedic provider are concrete. First, observed correlations alone are misleading, since Self-Efficacy appears to be the strongest lever in the data but disappears once confounders are controlled for. We find that the biggest gain will come from interventions designed to make individuals believe they are capable of performing recommended behaviour and to change the perceptions held by significant others (15%). We also find that interventions that integrate health concerns into a person's daily life and improve consumers' physical and mental well-being can lead to a substantial increase in the intention to consult an Ayurvedic doctor for diabetes (9.6%). Changing the perception held by significant others (society) about whether one should consult an Ayurvedic doctor for diabetes will also lead to a true causal lift of 7.5% from the baseline. However, it will require a systemic, long-term intervention to achieve the results.

## 6. Validation and sensitivity analysis

We examine the methodology for deducing CPTs from synthetic agents on two fronts. First, do the resulting probabilities encode plausible domain knowledge? Second, how robust are the decision-relevant outputs to choices made along the way?

### 6.1 Face validation through structural assumption testing

We treat the eleven structural assumptions as a form of face validation. Each assumption derives from the social cognition literature and predicts a directional change in $P(I = High)$ under a specified observation or intervention. If the LLM-elicited CPTs encode theory-consistent behavior, the assumptions should be validated under the network's belief propagation. The results support this expectation, where seven of the eleven assumptions were validated. The directional changes in posterior probability matched the predictions from the HBM, PMT and the TPB. For example, customers with high self-efficacy showed higher intention, whereas greater health consciousness was associated with higher intention. In contrast, higher perceived barriers to Ayurvedic care were associated with lower intention.

Three assumptions were rejected, and the rejections were interpretable rather than pathological. SA3a, that favorable attitude alone leads to high intention, fails because attitude does not operate in isolation. The full network suggests that attitude needs to combine with sufficient self-efficacy or supportive subjective norms to translate into intention. SA7a, that higher perceived barriers of the competing system favor Ayurveda, fails because the customer's attitude is shaped more by absolute properties of Ayurveda than by relative weaknesses of Western medicine.

The pass-and-fail pattern provides face validation in two senses. The model encodes most directional predictions from theory. The failures point to mediation effects that the network structure encodes, as well as the several paths through which belief propagates. Neither finding would be possible if the LLM-elicited CPTs were random or systematically biased.

### 6.2 Sensitivity to LLM parameters and virtual sample size

The methodology has three tuneable choices that may affect the resulting CPTs and downstream inference. These are the panel size N, the choice of LLM model, and the aggregation rule. Panel size N was set to 100 in the application, reflecting a cost-time balance. Smaller panels reduce costs but increase variance in aggregated CPT cells, while larger panels reduce variance at the cost of additional API calls and runtime. The total call count grows linearly with N. The aggregation rule was the trimmed mean with 10% trimming at the top and bottom. At N = 100, the trimmed mean uses 80 estimates per cell, which is sufficient to suppress individual-response noise. The trimmed mean implements the empirical finding of Doshi et al. (2024) that aggregated LLM responses resemble expert rankings even when individual responses are biased, and the trimming further reduces the effect of hallucinations and extreme outliers. A simple mean would weight outliers equally with the consensus, which we judge less robust. The model used was Gemini 2.5 Flash, reflecting the same cost-quality trade-off. Frontier-grade models such as GPT-4 or Claude Sonnet would deliver more nuanced responses at higher per-call cost, while smaller models would deliver lower-quality responses at lower cost.

### 6.3 Structural sensitivity

Structural sensitivity tests how decision recommendations change when the BBN structure changes. While a complete structural sensitivity analysis would require running the methodology on multiple alternative DAGs, an important form of structural testing is already embedded in the causal queries. The do-calculus implements interventions through graph mutilation. Each $do(X)$ operation severs all incoming edges to the intervention node $X$, which corresponds to a structural modification of the original network (Pearl, 2000). The truncated factorization computes $P(Y \mid do(X))$ on this mutilated graph (Pearl, 2000). Every causal query is therefore a sensitivity test of the original structure under a specific edge-removal scenario.

The eleven structural assumptions, therefore, double as a structural test battery. Seven assumptions failed to reject, with network behavior under intervention carrying the expected causal weight under graph mutilation. We provide the details in Appendix B, which show that seven SAs passed the causal test and four did not. These rejections do not constitute proof that the network structure is wrong. It suggests that such interventions may not be beneficial in the context under study. The face validation and the structural sensitivity captured by do-calculus interventions provide partial validation of the methodology. External benchmarking, systematic LLM-parameter experiments, and a complete alternative-structure analysis remain the open priorities, discussed as future work.

## 7. Discussion

### 7.1 Methodological Implications for OR

The methodology proposes a middle path between two long-standing approaches to BBN parameter learning. Pure expert elicitation captures tacit domain knowledge but incurs cognitive load and introduces subjectivity. Pure data-driven learning produces objective parameters but requires large representative datasets that management research rarely possesses. Our LLM-assisted approach occupies the space between these two endpoints.

The methodology aligns with the structure-learning categorization. We use an expert-based approach to structure, drawing on theory rather than data, thereby preserving causal interpretability and avoiding the limitations of correlation-based structure search. For parameter learning, we use a hybrid approach in which the LLM provides scalable elicitation while forward sampling and MLE refinement provide self-consistency. The combination gives the user CPTs grounded in domain context without the time cost of human surveys.

We operationalize the Bayesian decision models in domains where data scarcity has historically been a barrier lowering the entry cost for scholars and practitioners working in these areas. The integration with do-calculus enables causal queries that complement those offered by regression and structural equation modelling. The contrast between observational and interventional probabilities revealed is not directly available from those methods without strong assumptions. The methodology is reusable as the same workflow applies to any decision problem that can be framed as a BBN. It requires some adaptation of domain-specific persona descriptions and prompt templates.

We posit that the methodology complements rather than replaces human expertise. The user still defines the structure, writes the node descriptions, and reviews the CPTs. The LLM accelerates a single bottleneck step. Our approach brings synergy to human-AI collaboration as the way forward in contexts where data scarcity has been a problem.

### 7.2 Practical Considerations for Adoption

Several practical factors influence whether and how one adopts the methodology. The most prominent being cost, required expertise, interpretability, and governance. Cost scales linearly with panel size and the number of parent configurations. The application costs approximately USD 6 for a panel of 100 agents, and even larger panels with frontier-grade models rarely exceed a few hundred dollars. Some of our experiments, owing to a wrong set-up, cost upward of USD 200. However, compared to traditional human surveys, which take weeks and cost orders of magnitude more, LLM-assisted elicitation is fast and cheap.

The expertise required to adapt the methodology overlaps substantially with that of conventional BBN modelling. The user must specify a DAG, define discrete states, and write meaningful node descriptions. Additional skills in prompt design and a working understanding of LLM behavior can make the network more robust. The user defending the model can show exactly how each value was elicited, thus making the model interpretable. The trimmed-mean CPTs are interpretable in the same way that any BBN's CPTs are interpretable, and the DAG remains transparent. The origin of each parameter is documented through the persona description, the prompt template, and the cached responses. Organizations adopting LLM-assisted elicitation should record the LLM provider, model version, persona description, and prompt templates, and should establish review checkpoints. This will ground the governance aspect, which could be of concern while using an LLM-based survey.

The methodology is most attractive for early-stage decision modelling where the user needs an initial network quickly or in domains where human survey data are expensive or slow to collect. It is also useful to explore the problems where causal reasoning matters more than precise parameter values.

### 7.3 Limitations

The most fundamental limitation is the absence of full real-world validation. The simulation in Section 5 draws 100,000 samples from the LLM-elicited CPTs themselves and therefore cannot validate against actual customer behavior. The natural next step is to combine LLM priors with real data via Bayesian updating, where the LLM CPTs serve as priors and empirical observations update them through standard Bayesian inference. Replacing the LLM survey with an actual human survey for at least one decision problem would also benchmark LLM elicitation against ground truth. Agreement on direction and approximate magnitude would strengthen the case for routine elicitation. Any divergence would itself be informative for prompt design and aggregation.

The methodology depends heavily on the quality of the underlying LLM and on the prompt design. We assume the model has a reasonable representation of the decision context in its training data. This held for mainstream Indian customer behavior with Gemini 2.5 Flash but may break down for other geographies or niche populations. Surveying multiple LLM providers (Gemini,

Claude, GPT, and others) and aggregating the results to reach a consensus would mitigate provider-specific bias. Adding a prompt-sensitivity test before deployment in high-stakes decisions is a natural complement.

A panel size of N = 100 is sufficient for the trimmed mean to operate, but it is small relative to the population the synthetic agents represent. Larger panels would tighten estimates and allow stratification by demographic dimensions collapsed into the persona description. A structural limitation is that the user must specify the DAG before parameter elicitation, since the structure cannot be inferred from LLM responses. For domains where the structure is uncertain, the user can either commit to a theoretical structure or run the methodology across multiple candidates and compare them, as discussed in Section 6.3.

## 8. Conclusion

This paper proposes an LLM-assisted methodology for constructing Bayesian Belief Networks in the presence of data scarcity. The methodology addresses a long-standing bottleneck in BBN-based decision support. Expert elicitation places heavy cognitive demands on respondents and is hard to validate, while data-driven learning requires substantial volumes of representative data that are difficult to assemble. We position large language models as a third path. A panel of synthetic agents responds to structured prompts about each conditional probability cell, and a trimmed-mean aggregation rule reduces individual-response noise. The resulting CPTs initialize the BBN after which forward sampling and maximum likelihood estimation refine the parameters into a self-consistent network. This supports probabilistic and causal inference through truncated factorization. The methodology is delivered as a six-step workflow with a clean handoff between the LLM and the BBN engine. The user defines the structure and the discrete states, the LLM elicits the initial CPTs, and the BBN engine handles visualization, simulation, and inference. The user remains in the loop throughout for inspection and query specification.

We demonstrate the methodology on a health-services decision problem to understand the customer's intention to consult an Ayurvedic doctor for diabetes management. We rely on eight variables drawn from social cognition theory. The simulation reveals decision-relevant insights that observational analysis alone could not surface. Self-Efficacy looks like a strong lever in the data, but disappears once confounders are controlled, while Subjective Norms show suppressed correlation but a substantial true causal effect.

The contribution to operations research is threefold. We develop a scalable methodology for parameterizing Bayesian decision models when data is scarce. We also provide a working integration of generative AI into the BBN parameter-learning pipeline, through a reusable web-based application that applies to any decision problem framed as a BBN. Practitioners in supply chain risk, project management, manufacturing, and policy analysis can adopt the same workflow with domain-specific persona descriptions and prompt templates.

The simulation findings encourage a broader claim that aggregated LLM responses have begun to encode meaningful slices of human reasoning about uncertain decisions. The trimmed mean across a diverse synthetic panel produces CPTs that reproduce seven out of eleven theoretical predictions. It surfaces interpretable rejections for the rest, suggesting that traditional machine learning and generative AI can coexist and augment each other in advancing decision science. Human AI collaboration can shape the future of decision sciences research. The next phase of research should validate the methodology against empirical surveys and extend it to other operations domains.

Declaration of interests:

The authors declare that they have no known competing financial interests or personal relationships that could have appeared to influence the work reported in this paper.

Funding sources: This research did not receive any specific grant from funding agencies in the public, commercial, or not-for-profit sectors

## Appendix A: The Learnt CPT Table

*Learnt CPT for PBa, PBaC and HS Node*

| **Node: PBa** | | **Node: PBaC** | | **Node: HS** | |
|---|---|---|---|---|---|
| State | Probability | State | Probability | State | Probability |
| Low | 0.3942 | Low | 0.2997 | Bad | 0.5287 |
| Medium | 0.3578 | Medium | 0.3754 | Good | 0.4713 |
| High | 0.2481 | High | 0.3249 | | |

*Learnt CPT for A and SE Node*

| Node: A | | | | | Node: SE | | | |
|---|---|---|---|---|---|---|---|---|
| HS | PBa | PBaC | State | Probability | A | HC | State | Probability |
| Bad | Low | Low | Low | 0.9518 | Low | Low | Low Control | 0.8363 |
| Bad | Low | Low | High | 0.0482 | Low | Low | Average Control | 0.1168 |
| Bad | Low | Medium | Low | 0.8853 | Low | Low | High Control | 0.0469 |
| Bad | Low | Medium | High | 0.1147 | Low | Moderate | Low Control | 0.6752 |
| Bad | Low | High | Low | 0.8269 | Low | Moderate | Average Control | 0.2589 |
| Bad | Low | High | High | 0.1731 | Low | Moderate | High Control | 0.0659 |
| Bad | Medium | Low | Low | 0.7875 | Low | High | Low Control | 0.5407 |
| Bad | Medium | Low | High | 0.2125 | Low | High | Average Control | 0.3533 |
| Bad | Medium | Medium | Low | 0.7619 | Low | High | High Control | 0.106 |
| Bad | Medium | Medium | High | 0.2381 | High | Low | Low Control | 0.2633 |
| Bad | Medium | High | Low | 0.492 | High | Low | Average Control | 0.4087 |
| Bad | Medium | High | High | 0.508 | High | Low | High Control | 0.328 |
| Bad | High | Low | Low | 0.3619 | High | Moderate | Low Control | 0.0594 |
| Bad | High | Low | High | 0.6381 | High | Moderate | Average Control | 0.2569 |
| Bad | High | Medium | Low | 0.2942 | High | Moderate | High Control | 0.6838 |
| Bad | High | Medium | High | 0.7058 | High | High | Low Control | 0.0449 |
| Bad | High | High | Low | 0.2767 | High | High | Average Control | 0.1401 |
| Bad | High | High | High | 0.7233 | High | High | High Control | 0.815 |
| Good | Low | Low | Low | 0.876 | | | | |
| Good | Low | Low | High | 0.124 | | | | |
| Good | Low | Medium | Low | 0.7723 | | | | |
| Good | Low | Medium | High | 0.2277 | | | | |
| Good | Low | High | Low | 0.6316 | | | | |
| Good | Low | High | High | 0.3684 | | | | |
| Good | Medium | Low | Low | 0.6119 | | | | |
| Good | Medium | Low | High | 0.3881 | | | | |
| Good | Medium | Medium | Low | 0.4883 | | | | |
| Good | Medium | Medium | High | 0.5117 | | | | |
| Good | Medium | High | Low | 0.3233 | | | | |
| Good | Medium | High | High | 0.6767 | | | | |
| Good | High | Low | Low | 0.1723 | | | | |
| Good | High | Low | High | 0.8277 | | | | |
| Good | High | Medium | Low | 0.1144 | | | | |
| Good | High | Medium | High | 0.8856 | | | | |
| Good | High | High | Low | 0.0936 | | | | |
| Good | High | High | High | 0.9064 | | | | |

*Learnt CPT for HC and SN Node*

| **Node: HC** | | | **Node: SN** | | |
|---|---|---|---|---|---|
| HS | State | Probability | SE | State | Probability |
| Bad | Low | 0.0602 | Low Control | Not Influenced | 0.1026 |
| Bad | Moderate | 0.2307 | Low Control | Somewhat Influenced | 0.3201 |
| Bad | High | 0.7091 | Low Control | Highly Influenced | 0.5773 |
| Good | Low | 0.0721 | Average Control | Not Influenced | 0.3081 |
| Good | Moderate | 0.2617 | Average Control | Somewhat Influenced | 0.5036 |
| Good | High | 0.6661 | Average Control | Highly Influenced | 0.1883 |
| | | | High Control | Not Influenced | 0.7278 |
| | | | High Control | Somewhat Influenced | 0.1933 |
| | | | High Control | Highly Influenced | 0.0789 |

*Learnt CPT for Intention to consult an Ayurvedic doctor Node*

| **Node: I** |
|---|

| A | HC | SE | SN | State | Probability |
|---|---|---|---|---|---|
| Low | Low | Low Control | Not Influenced | Low | 0.9717 |
| Low | Low | Low Control | Not Influenced | High | 0.0283 |
| Low | Low | Low Control | Somewhat Influenced | Low | 0.927 |
| Low | Low | Low Control | Somewhat Influenced | High | 0.073 |
| Low | Low | Low Control | Highly Influenced | Low | 0.8659 |
| Low | Low | Low Control | Highly Influenced | High | 0.1341 |
| Low | Low | Average Control | Not Influenced | Low | 0.9286 |
| Low | Low | Average Control | Not Influenced | High | 0.0714 |
| Low | Low | Average Control | Somewhat Influenced | Low | 0.8938 |
| Low | Low | Average Control | Somewhat Influenced | High | 0.1062 |
| Low | Low | Average Control | Highly Influenced | Low | 0.7375 |
| Low | Low | Average Control | Highly Influenced | High | 0.2625 |
| Low | Low | High Control | Not Influenced | Low | 0.9076 |
| Low | Low | High Control | Not Influenced | High | 0.0924 |
| Low | Low | High Control | Somewhat Influenced | Low | 0.8537 |
| Low | Low | High Control | Somewhat Influenced | High | 0.1463 |
| Low | Low | High Control | Highly Influenced | Low | 0.7368 |
| Low | Low | High Control | Highly Influenced | High | 0.2632 |
| Low | Moderate | Low Control | Not Influenced | Low | 0.9238 |
| Low | Moderate | Low Control | Not Influenced | High | 0.0762 |
| Low | Moderate | Low Control | Somewhat Influenced | Low | 0.8711 |
| Low | Moderate | Low Control | Somewhat Influenced | High | 0.1289 |
| Low | Moderate | Low Control | Highly Influenced | Low | 0.7298 |
| Low | Moderate | Low Control | Highly Influenced | High | 0.2702 |
| Low | Moderate | Average Control | Not Influenced | Low | 0.8588 |
| Low | Moderate | Average Control | Not Influenced | High | 0.1412 |
| Low | Moderate | Average Control | Somewhat Influenced | Low | 0.7965 |
| Low | Moderate | Average Control | Somewhat Influenced | High | 0.2035 |
| Low | Moderate | Average Control | Highly Influenced | Low | 0.5839 |
| Low | Moderate | Average Control | Highly Influenced | High | 0.4161 |
| Low | Moderate | High Control | Not Influenced | Low | 0.807 |
| Low | Moderate | High Control | Not Influenced | High | 0.193 |
| Low | Moderate | High Control | Somewhat Influenced | Low | 0.7513 |
| Low | Moderate | High Control | Somewhat Influenced | High | 0.2487 |
| Low | Moderate | High Control | Highly Influenced | Low | 0.5224 |
| Low | Moderate | High Control | Highly Influenced | High | 0.4776 |
| Low | High | Low Control | Not Influenced | Low | 0.8287 |
| Low | High | Low Control | Not Influenced | High | 0.1713 |

| Low | High | Low Control | Somewhat Influenced | Low | 0.7463 |
|---|---|---|---|---|---|
| Low | High | Low Control | Somewhat Influenced | High | 0.2537 |
| Low | High | Low Control | Highly Influenced | Low | 0.6621 |
| Low | High | Low Control | Highly Influenced | High | 0.3379 |
| Low | High | Average Control | Not Influenced | Low | 0.7057 |
| Low | High | Average Control | Not Influenced | High | 0.2943 |
| Low | High | Average Control | Somewhat Influenced | Low | 0.6244 |
| Low | High | Average Control | Somewhat Influenced | High | 0.3756 |
| Low | High | Average Control | Highly Influenced | Low | 0.4638 |
| Low | High | Average Control | Highly Influenced | High | 0.5362 |
| Low | High | High Control | Not Influenced | Low | 0.7322 |
| Low | High | High Control | Not Influenced | High | 0.2678 |
| Low | High | High Control | Somewhat Influenced | Low | 0.5681 |
| Low | High | High Control | Somewhat Influenced | High | 0.4319 |
| Low | High | High Control | Highly Influenced | Low | 0.4606 |
| Low | High | High Control | Highly Influenced | High | 0.5394 |
| High | Low | Low Control | Not Influenced | Low | 0.6747 |
| High | Low | Low Control | Not Influenced | High | 0.3253 |
| High | Low | Low Control | Somewhat Influenced | Low | 0.6043 |
| High | Low | Low Control | Somewhat Influenced | High | 0.3957 |
| High | Low | Low Control | Highly Influenced | Low | 0.5205 |
| High | Low | Low Control | Highly Influenced | High | 0.4795 |
| High | Low | Average Control | Not Influenced | Low | 0.5237 |
| High | Low | Average Control | Not Influenced | High | 0.4763 |
| High | Low | Average Control | Somewhat Influenced | Low | 0.4731 |
| High | Low | Average Control | Somewhat Influenced | High | 0.5269 |
| High | Low | Average Control | Highly Influenced | Low | 0.3395 |
| High | Low | Average Control | Highly Influenced | High | 0.6605 |
| High | Low | High Control | Not Influenced | Low | 0.4422 |
| High | Low | High Control | Not Influenced | High | 0.5578 |
| High | Low | High Control | Somewhat Influenced | Low | 0.3464 |
| High | Low | High Control | Somewhat Influenced | High | 0.6536 |
| High | Low | High Control | Highly Influenced | Low | 0.2405 |
| High | Low | High Control | Highly Influenced | High | 0.7595 |
| High | Moderate | Low Control | Not Influenced | Low | 0.5 |
| High | Moderate | Low Control | Not Influenced | High | 0.5 |
| High | Moderate | Low Control | Somewhat Influenced | Low | 0.4327 |
| High | Moderate | Low Control | Somewhat Influenced | High | 0.5673 |
| High | Moderate | Low Control | Highly Influenced | Low | 0.3097 |

| High | Moderate | Low Control | Highly Influenced | High | 0.6903 |
|---|---|---|---|---|---|
| High | Moderate | Average Control | Not Influenced | Low | 0.1418 |
| High | Moderate | Average Control | Not Influenced | High | 0.8582 |
| High | Moderate | Average Control | Somewhat Influenced | Low | 0.128 |
| High | Moderate | Average Control | Somewhat Influenced | High | 0.872 |
| High | Moderate | Average Control | Highly Influenced | Low | 0.083 |
| High | Moderate | Average Control | Highly Influenced | High | 0.917 |
| High | Moderate | High Control | Not Influenced | Low | 0.0921 |
| High | Moderate | High Control | Not Influenced | High | 0.9079 |
| High | Moderate | High Control | Somewhat Influenced | Low | 0.0851 |
| High | Moderate | High Control | Somewhat Influenced | High | 0.9149 |
| High | Moderate | High Control | Highly Influenced | Low | 0.0586 |
| High | Moderate | High Control | Highly Influenced | High | 0.9414 |
| High | High | Low Control | Not Influenced | Low | 0.4043 |
| High | High | Low Control | Not Influenced | High | 0.5957 |
| High | High | Low Control | Somewhat Influenced | Low | 0.3181 |
| High | High | Low Control | Somewhat Influenced | High | 0.6819 |
| High | High | Low Control | Highly Influenced | Low | 0.174 |
| High | High | Low Control | Highly Influenced | High | 0.826 |
| High | High | Average Control | Not Influenced | Low | 0.11 |
| High | High | Average Control | Not Influenced | High | 0.89 |
| High | High | Average Control | Somewhat Influenced | Low | 0.1017 |
| High | High | Average Control | Somewhat Influenced | High | 0.8983 |
| High | High | Average Control | Highly Influenced | Low | 0.0654 |
| High | High | Average Control | Highly Influenced | High | 0.9346 |
| High | High | High Control | Not Influenced | Low | 0.0506 |
| High | High | High Control | Not Influenced | High | 0.9494 |
| High | High | High Control | Somewhat Influenced | Low | 0.0537 |
| High | High | High Control | Somewhat Influenced | High | 0.9463 |
| High | High | High Control | Highly Influenced | Low | 0.0238 |
| High | High | High Control | Highly Influenced | High | 0.9762 |